\documentclass[final]{cvpr}

\usepackage{times}
\usepackage{epsfig}
\usepackage{graphicx}
\usepackage{amsmath}
\usepackage{amssymb}

\usepackage{multirow}
\usepackage{diagbox}

\def\eg{\emph{e.g}\onedot} 
\def\ie{\emph{i.e}\onedot} 
 
\def\etc{\emph{etc}\onedot} 
 
\def\etal{\emph{et al}\onedot}

\newlength\savewidth

\usepackage{array}
\usepackage[table]{xcolor}
\usepackage{colortbl}
\definecolor{bblue}{rgb}{0,150,230}
\definecolor{mygray}{gray}{.9}
\definecolor{myy}{RGB}{126,95,0}
\definecolor{mygreen}{RGB}{93,174,86}
\usepackage{bm}

\newcommand\blfootnote[1]{%
  \begingroup
  \renewcommand\thefootnote{}\footnote{#1}%
  \addtocounter{footnote}{-1}%
  \endgroup
}

\newcolumntype{I}{!{\vrule width 1pt}}

\definecolor{ggray}{RGB}{127,127,127}

\makeatletter
\newcommand{\thickhline}{%
    \noalign {\ifnum 0=`}\fi \hrule height 1pt
    \futurelet \reserved@a \@xhline
}
\makeatother

\usepackage{caption}
\usepackage[pagebackref=true,breaklinks=true,colorlinks,bookmarks=false]{hyperref}
\def\confYear{CVPR 2021}

\usepackage{cleveref}
\crefname{section}{§}{§§}
\Crefname{section}{§}{§§}

\newcommand*{\affaddr}[1]{#1}
\newcommand*{\affmark}[1][*]{\textsuperscript{#1}}

\begin{document}

\title{Rethinking Cross-modal Interaction from a Top-down Perspective \\
for Referring Video Object Segmentation}

\author{
Chen Liang\affmark[1,3], Yu Wu\affmark[3,4], Tianfei Zhou\affmark[2], Wenguan Wang\affmark[2], Zongxin Yang\affmark[3,4], Yunchao Wei\affmark[4] and Yi Yang\affmark[1]\\
\small
\affaddr{\affmark[1]Zhejiang University} \quad
\affaddr{\affmark[2]ETH Zurich} \quad
\affaddr{\affmark[3]Baidu Research} \quad
\affaddr{\affmark[4]University of Technology Sydney}\\
}

\maketitle

\begin{abstract}
	\blfootnote{Extended version published in~\cite{liang2023locater}.} Referring video object segmentation (RVOS) aims to segment video objects with the guidance of natural language reference.
	Previous methods typically tackle RVOS through directly grounding linguistic reference over the image lattice. Such bottom-up strategy fails to explore object-level cues, easily leading to inferior results. In this work, we instead put forward a two-stage, top-down RVOS solution. First, an exhaustive set of object tracklets is constructed by propagating object masks detected from several sampled frames to the entire video. Second, a Transformer-based tracklet-language grounding module is proposed, which models instance-level visual relations and cross-modal interactions simultaneously and efficiently. 	Our model ranks $1^{st}$ place on CVPR2021 Referring Youtube-VOS challenge.

\end{abstract}

\vspace{-4pt}
\section{Introduction}
\vspace{-2pt}

\begin{figure}[t]
	\begin{center}
		\includegraphics[width=1\linewidth]{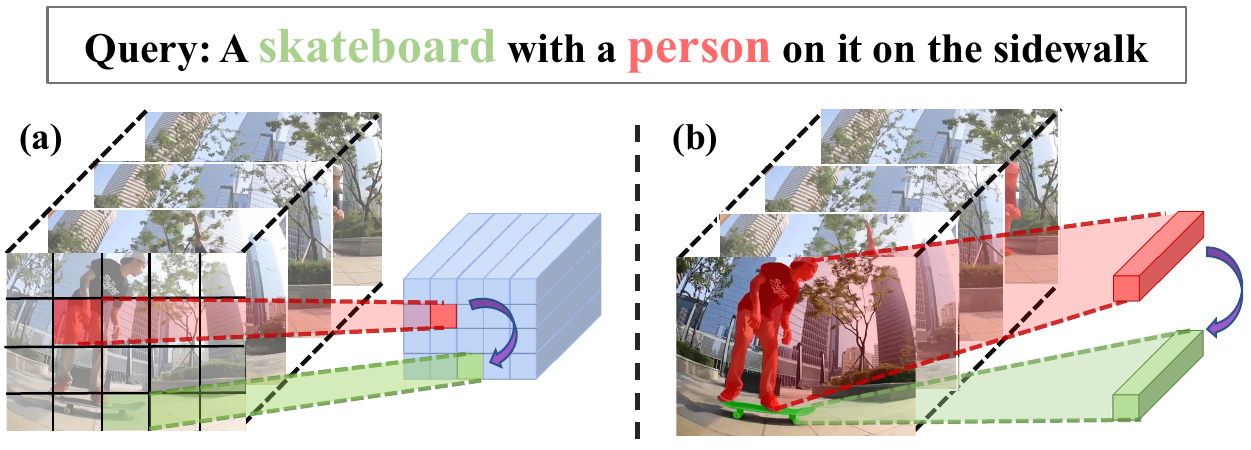}
	\end{center}
	\vspace{-12pt}
	\captionsetup{font=small}
	\caption{\small\textbf{An illustration of our motivation}.
		Previous bottom-up methods \textbf{(a)} perform cross-modal interaction at grid level, and fail to capture crucial object-level relations as top-down approach \textbf{(b)}.
	}\label{fig:moti}
	\vspace{-14pt}
\end{figure}

Referring video object segmentation (RVOS) targets at segmenting video objects referred by given language expressions.  RVOS is a challenging task as it requires not only comprehensive understanding the semantics within individual modalities, but also pixel-level cross-modal reasoning. 
Existing RVOS models~\cite{seo2020urvos,khoreva2018video} typically work in a bottom-up fashion (Fig.~\ref{fig:moti}-(a)), \ie, perform grid-level alignment between visual and linguistic modalities. Thus they lack explicit knowledge about visual objects, leading to unreliable cross-modal reasoning and inaccurate segmentation.

In this work, we rethink RVOS from a top-down perspective (Fig.~\ref{fig:moti}-(b)), by comprehensively exploring cross-object relations and conducting object-level cross-modal grounding.  With a similar spirit of~\cite{liang2021clawcranenet}, our approach mainly consists of two stages: object tracklet generation and tracklet-language grounding. In the first stage, we generate a set of high-quality object tracklets from input videos. Then, in the second stage, we ground the reference over the detected tracklets and select the best-matched one as the final output.

More specifically, in the object tracklet generation stage, a lot of object candidate masks are first generated by applying instance segmentation over several sampled key frames. We further propagate the detected candidate masks to the whole video sequence, and generate an exhaustive set of object tracklets. After that, a tracklet-NMS mechanism is designed to remove redundant tracklets and select the high-quality ones as candidates for language-guided segmentation. In the tracklet-language grounding stage, we build a Transformer-based grounding module. Benefiting from the powerful self-attention computation within the Transformer blocks, the within-modal relations among objects and inter-modal interactions between tracklets and language can be comprehensively and efficiently modeled.

Our model ranked $1^{st}$ place in $3^{rd}$ Large-scale Video Object Segmentation Challenge (CVPR2021): Referring Video Object Segmentation track~\cite{vosc2021}, with an overall $\mathcal{J}$\&$\mathcal{F}$ of $61.4\%$ and $60.7\%$ on \texttt{test}-\texttt{dev} and \texttt{test}-\texttt{challenge}, respectively.

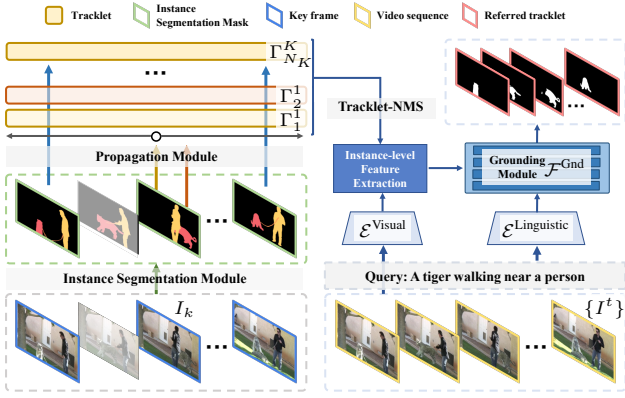
\begin{figure}[t]
	\begin{center}
		\includegraphics[width=1\linewidth]{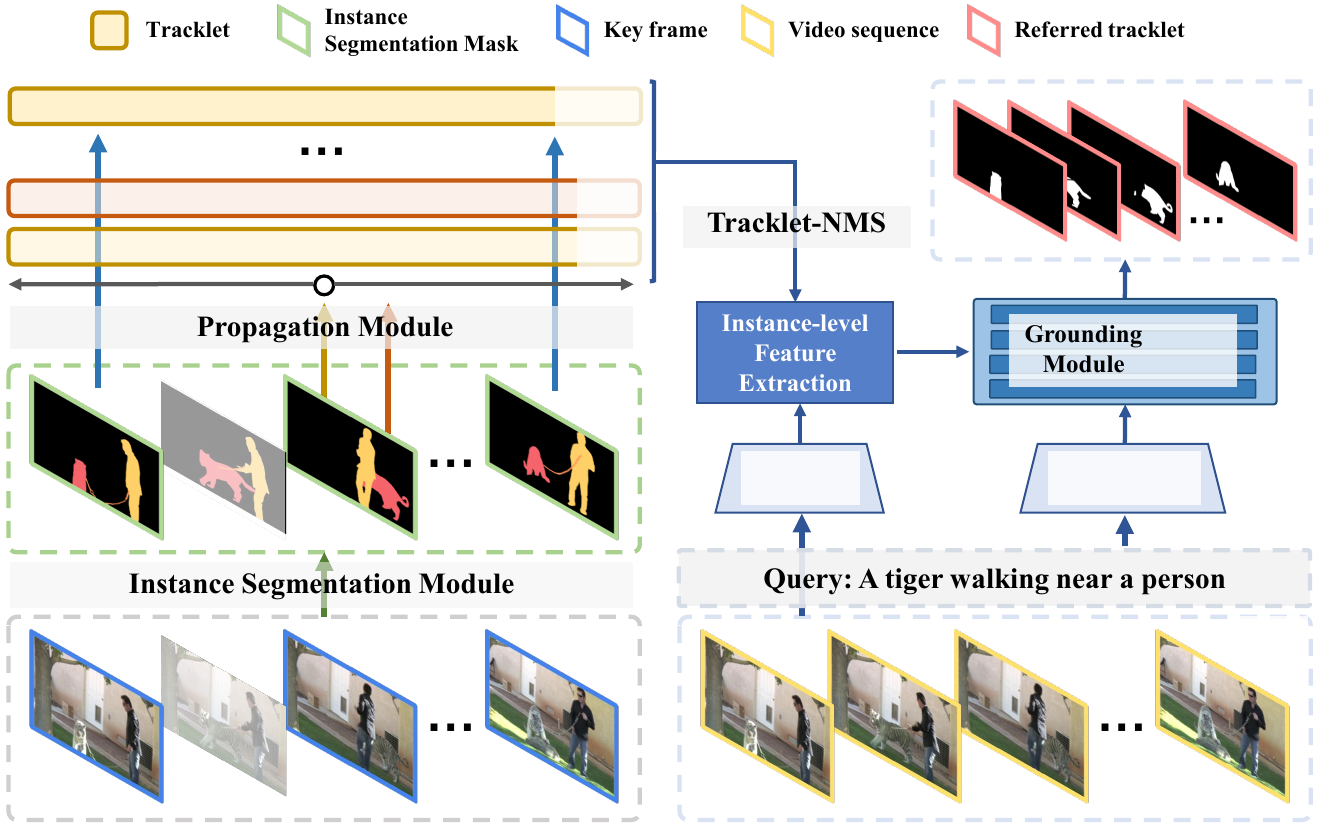}
		\put(-102, 60){\scriptsize$\mathcal{E}^\text{Visual}$}
		\put(-48, 60){\scriptsize$\mathcal{E}^\text{Linguistic}$}
		\put(-32, 83){\scriptsize$\mathcal{F}^\text{Gnd}$}
		\put(-132, 102){\scriptsize$\Gamma^1_1$}
		\put(-132, 111){\scriptsize$\Gamma^1_2$}
		\put(-136, 128){\scriptsize$\Gamma^K_{N_K}$}
		\put(-172, 31){\scriptsize$I_k$}
		\put(-17, 31){\scriptsize$\{I^t\}$}
	\end{center}
	\vspace{-12pt}
	\captionsetup{font=small}
	\caption{\small\textbf{Pipeline of our proposed method}, which contains two major stages, \ie, object tracklet generation (left column) and tracklet-language grounding (right column).
	}\label{fig:pipe}
	\vspace{-14pt}
\end{figure}

\vspace{-2pt}
\section{Related Work}
\vspace{-1pt}
\noindent\textbf{RVOS Datasets.} The task of RVOS is proposed in~\cite{khoreva2018video}, which mainly focuses on actor behavior understanding within an limited predefined action categories. Recently, Seo \etal introduced a new large-scale dataset~\cite{seo2020urvos}, \ie, Refer-Youtube-VOS (RVOS-D), derived from Youtube-VOS~\cite{xu2018youtube}. RVOS-D provides more complex language descriptions from a broader object categories ($\sim$\!~90 categories) within relatively longer video sequences ($\sim$\!~6\!~s). Thus it poses more challenges for RVOS methods.  Referring Youtube-VOS challenge~\cite{vosc2021} is built upon RVOS-D~\cite{seo2020urvos}.

\noindent\textbf{RVOS Methods.} Current studies in the filed of RVOS are made mainly around the theme of building effective multi-modal feature representations. Existing methods typically make use of dynamic convolutions~\cite{wang2020context,gavrilyuk2018actor} to adaptively generate convolutional filters that better respond to the referent, or leverage cross-modal attention~\cite{wang2019asymmetric,ningpolar} to compute the correlations among input visual and  linguistic embeddings. However, these methods only approach RVOS on the grid level, ignoring the importance of object-level visual cues.

\vspace{-1pt}
\section{Methodology}
\vspace{-1pt}

\noindent\textbf{Overview.}
Given a video sequence $\mathcal{I}\!=\!\{I^{t\!}\!\in\!\mathbb{R}^{W \times H \times 3} \}_{t=1}^{T}$ with $T$ frames, and corresponding referring expression $Q\!=\!\{q^l\}_{l=1}^{L}$ with $L$ words, a set of referred object segmentation masks $\{S^{t\!}\!\in\!\{0,1\}^{W \times H} \}_{t=1}^{T}$ is requested for RVOS.
As illustrated in Fig.~\ref{fig:pipe}, we design a two-down approach with two stages for object tracklet generation and tracklet-language grounding, respectively. In the first stage, we construct a comprehensive set of object candidate tracklets from $\mathcal{I}$ and propose a sequence-nms module for reducing redundant candidates. In the second stage, the referred target is selected from the tracklets under the guidance of $Q$.

\noindent\textbf{Object Tracklet Construction.}
We first uniformly sample $K$ frames from $\mathcal{I}$. For each key frame $I^k$, a set of mask candidates, \ie, $\mathcal{O}^k$, are generated through an image instance segmentation model $\mathcal{F}^{\text{Seg}}$:
\begin{equation}\small
	\begin{aligned}\label{eq:0}
		\mathcal{O}^k = \{O^k_n\}_{n=1}^{N_k}= \mathcal{F}^{\text{Seg}}(I^k),
	\end{aligned}
\end{equation}
where $N_k$ refers to the number of the candidates in $I^k$, and, for each mask candidate $O^k_n\!\in\!\mathcal{O}^k$, we have $O^k_n\!\in\!\{0,1\}^{W \times H}$.
Then, a video mask propagation model $\mathcal{F}^{\text{Prop}}$ is applied for each $O^k_n\!\in\!\mathcal{O}^k$, to forward and backward propagate the mask to the entire video and get  corresponding object tracklet $\Gamma^k_n$:
\begin{equation}\small
	\begin{aligned}\label{eq:1}
		\Gamma^k_n= \mathcal{F}^{\text{Prop}}(O^k_n, \mathcal{I})\!\in\!\{0,1\}^{T \times W \times H}.
	\end{aligned}
\end{equation}
Thus each tracklet is a sequence of masks, \ie, $\Gamma^k_n\!=\!\{M^{k,t}_{n}\!\in\!\{0,1\}^{W \times H}\}_{t=1}^{T}$, corresponds to the object candidate $O^k_n$ in key frame $I^k$. And we define $\mathcal{T}^k = \{\Gamma^k_n\}_{n=1}^{N_k}$  as the set of all the tracklets generated from $\mathcal{O}^k$.

Based on above strategy, we generate a lot of tracklets, \ie, $\{\mathcal{T}^k\}_k$, from the $K$ key frames. This ensures that we can generate a complete object candidate set that covers object instances in $\mathcal{I}$ as many as possible, without the disturbance from object occlusion, and move-in/-out. We denote the set of all generated candidate tracklets as $\mathcal{T}\!=\!\cup_{k\in\{1\cdots K\}}\mathcal{T}^k$.

\noindent\textbf{Tracklet-NMS.} As we sample several key frames, there exist a lot of similar tracklets that correspond to the same object instance. This would bring an extra challenge to the following tracklet-language grounding process. Inspired by~\cite{lin2021video}, we introduce a tracklet-level NMS process that eliminates redundant candidates in $\mathcal{T}$ efficiently. We first define tracklet-IoU that measures the similarity  between two tracklets, \ie, $\Gamma_p, \Gamma_q\!\in\!\mathcal{T}$:
\begin{equation}\small
	\begin{aligned}\label{eq:2}
		\text{tracklet-IoU}(\Gamma_p, \Gamma_q) &= \frac{\sum_{t=1}^{T}| M^t_p \cap M^t_q|}{\sum_{t=1}^{T}| M^t_p \cup M^t_q|}, \\
	\end{aligned}
\end{equation}
where $\Gamma_p\!=\!\{M^t_q\}_{t=1}^{T}$, and $\Gamma_q\!=\!\{M^t_q\}_{t=1}^{T}$. Each tracklet $\Gamma$ is also assigned with a score, defined as the product of the confidence score of $O^k_n$ (obtained from $\mathcal{F}^{\text{Seg}}$) and the mask propagation probability (obtained from $\mathcal{F}^{\text{Prop}}$), averaged over all the $T$ frames. Based on the tracklet score and tracklet-IoU, traditional NMS algorithms~\cite{girshick2015fast,girshick2014rich} is conducted. As at most 5 objects might be requested in our concerned experimental setting, we keep at most $P\!=\!10$ tracklets with highest scores for each video after NMS. We refer the final tracklet set as $\hat{\mathcal{T}}\!=\!\{\Gamma_p\}_{p=1}^P$.

\begin{figure*}[t]
	\begin{center}
		\includegraphics[width=1\linewidth]{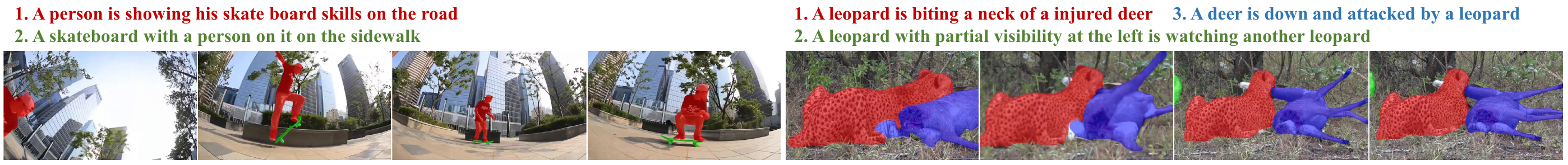}
	\end{center}
	\vspace{-15pt}
	\captionsetup{font=small}
	\caption{\small
		\textbf{Representative visual results} on RVOS-D \texttt{test-challenge} set. Each referent and the corresponding textual description are highlighted in the same color.
	}\label{fig:quali}
	\vspace{-10pt}
\end{figure*}

\noindent\textbf{Tracklet-Language Grounding.} We adopt per-frame reference grounding to determine the referred object from $\hat{\mathcal{T}}$. Each frame $I^t$ and linguistic input $Q$ are first fed into single-modality encoders, \ie,  separately for within-modal feature extraction:
\begin{equation}\small
	\begin{aligned}\label{eq:3}
		\bm{I}^t &= \mathcal{E}^{\text{Visual}}(I^t)~ \in \mathbb{R}^{w \times h \times D}, \\
		\bm{Q} &= \mathcal{E}^{\text{Linguistic}}(Q)~ \in \mathbb{R}^{L\times D}, \\
	\end{aligned}
\end{equation}
where $\bm{I}^t$ and $\bm{Q}$ are extracted visual and linguistic features, respectively.

For each tracklet $\Gamma_p\!\in\!\hat{\mathcal{T}}$, we extracted its corresponding feature at frame $I^t$ through:
\begin{equation}\small
	\begin{aligned}\label{eq:4}
		\bm{\tau}_p^t &= \text{AvgPool}(\bm{I}^t \otimes M_p^t)~ \in \mathbb{R}^{D},
	\end{aligned}
\end{equation}
where $M_p^t\!\in\!\{0,1\}^{W \times H}$ refers to the candidate mask of tracklet $\Gamma_p$ in frame $\bm{I}^t$, and $\otimes$ denotes hadamard product. Note that the rescaling process for feature dimension alignment is omitted.

Given concatenated embeddings $\bm{\Gamma}^t\!=\![\bm{\tau}_1^t,\cdots,\bm{\tau}_P^t]\!\in\!\mathbb{R}^{P \times D}$ for the candidate tracklets $\hat{\mathcal{T}}\!=\!\{\Gamma_p\}_{p=1}^P$ at frame $I^t$, and the linguistic representation $\bm{Q}$, we propose a Transformer-based~\cite{vaswani2017attention} grounding module $\mathcal{F}^{\text{Gnd}}$ for tracklet-language grounding:
\begin{equation}\small
	\begin{aligned}\label{eq:5}
		\{s^t_p\}^P_{p=1}= \mathcal{F}^{\text{Gnd}}([\bm{\Gamma}^t + \bm{e}^v, \bm{Q} + \bm{e}^l],\\
	\end{aligned}
\end{equation}
where $s^t_p\!\in\![0,1]$ is the grounding score of tracklet $\Gamma_p$ in $\bm{I}^t$.
Here $\bm{e}^v\!\in\!\mathbb{R}^{P\!\times\!D}$ and $\bm{e}^l\!\in\!\mathbb{R}^{L\!\times\!D}$ are learnable modal embeddings.
Due to the self-attention mechanism in the Transformer, the interactions among different object tracklets and between different modalities are comprehensively captured, leading to promising grounding performance.

The final grounding score for each $\Gamma_p$ is given as: $s_p\!=\!\text{Mean}(s^1_p,\cdots, s^T_p)$. The the segments $\{S^{t\!}\}_{t=1}^{T}$ are $\{M^t_{p*}\}_{t=1}^{T}$, where $p*\!=\!\arg\max_p(s_1,\cdots, s_P)$.


\vspace{-1pt}
\section{Experiment}
\vspace{-1pt}

\begin{table}[t]
	\centering
	\small
	\resizebox{0.49\textwidth}{!}{
		\setlength\tabcolsep{8pt}
		\begin{tabular}{r||ccc}
			\hline\thickhline
			\rowcolor{mygray}
			\textbf{Team} & $\mathcal{J}$\&$\mathcal{F}\uparrow$ & $\mathcal{J}\uparrow$ & $\mathcal{F}\uparrow$\\
			\hline \hline
			\textbf{leonnnop (Ours)} & \textbf{61.4} \color{mygreen}{(\textbf{+6.6})} & \textbf{60.0} \color{mygreen}{(\textbf{+6.3})} & \textbf{62.7} \color{mygreen}{(\textbf{+6.7})} \\
			\hline
			nowherespyfly & 54.8 & 53.7 & 56.0 \\
			seonguk & 48.9 & 47.0 & 50.8 \\
			wangluting & 48.5  & 47.1 & 49.9   \\
			Merci1 & 44.9 & 43.9 & 45.9   \\
			\hline
		\end{tabular}
	}
	\captionsetup{font=small}
	\caption{\small\textbf{Benchmarking results} on the \texttt{test-dev} set of Referring Youtube-VOS challenge.
	}
	\vspace{-14pt}
	\label{table:dev}
\end{table}

\noindent\textbf{Challenge Dataset and Evaluation Metrics.} We test our model on Referring Youtube-VOS challenge~\cite{vosc2021}, which is built upon the recently released RVOS-D dataset~\cite{seo2020urvos}.
The challenge dataset has $3,\!978$ videos with about 15K language reference sentences in total; $3,\!471$ videos are released with annotations for training. The rest videos are split into $202$/$305$  for constructing \texttt{test-dev}/\texttt{test-challenge} sets, whose annotations are preserved for benchmarking.
We use standard metrics, \ie, region similarity $\mathcal{J}$ and contour accuracy $\mathcal{F}$, for evaluation.

\noindent\textbf{Detailed Network Architecture.}
We employ two instance segmentation models, \ie, HTC~\cite{chen2019hybrid} and CondInst~\cite{tian2020conditional}, for implementing $\mathcal{F}^{\text{Seg}}$ in Eq.~\ref{eq:0}. At each key frame $I^k$, the mask candidates $\mathcal{O}_k$ are a combination of all proposals generated from the two models. The mask propagation model $\mathcal{F}^{\text{Prop}}$ (Eq.~\ref{eq:1}) is implemented as  CFBI+~\cite{yang2020collaborative}.
We uniformly sample $K\!=\!7$ key frames for each video sequence.
For tracklet-language grounding, we implement the visual encoder $\mathcal{E}^\text{Visual}$ as ResNet-101~\cite{he2016deep}  initialized from ImageNet-pretrained weights and linguistic encoder $\mathcal{E}^\text{Linguistic}$ as a standard $\text{BERT}_\text{BASE}$ model. The grounding module $\mathcal{F}^{\text{Gnd}}$ is  a 4-layer Transformer~\cite{vaswani2017attention} with 12 heads in each layer, followed by a 2-layer MLP and a softmax layer for probability prediction.
Input sentences are split by the WordPiece tokenizer~\cite{wu2016google} as in~\cite{devlin2018bert}.
Both the hidden dimensions of Transformer and feature channel of within-modal representations are set to $768$, \ie, $D\!=\!768$.

\begin{table}[t]
	\centering
	\small
	\resizebox{0.49\textwidth}{!}{
		\setlength\tabcolsep{8pt}
		\begin{tabular}{r||ccc}
			\hline\thickhline
			\rowcolor{mygray}
			\textbf{Team} & $\mathcal{J}$\&$\mathcal{F}\uparrow$ & $\mathcal{J}\uparrow$ & $\mathcal{F}\uparrow$\\
			\hline \hline
			\textbf{leonnnop (Ours)} & \textbf{60.7} \color{mygreen}{(\textbf{+11.3})}  & \textbf{59.4} \color{mygreen}{(\textbf{+11.0})} & \textbf{62.0} \color{mygreen}{(\textbf{+11.7})} \\
			\hline
			nowherespyfly & 49.4 & 48.4 & 50.3 \\
			feng915912132 & 48.2 & 47.4 & 49.0 \\
			Merci1 & 41.2 & 40.6 & 41.8   \\
			wangluting & 40.7 & 39.5 & 41.8   \\
			\hline
		\end{tabular}
	}
	\captionsetup{font=small}
	\caption{\small\textbf{Benchmarking results} on \texttt{test-challenge} set of Referring Youtube-VOS challenge.
	}
	\vspace{-8pt}
	\label{table:challenge}
\end{table}

\begin{table}[t]
	\centering
	\small
	\resizebox{0.49\textwidth}{!}{
		\setlength\tabcolsep{4pt}
		\begin{tabular}{l||ccc}
			\hline\thickhline
			\rowcolor{mygray}
			\textbf{Model} & $\mathcal{J}$\&$\mathcal{F}\uparrow$ & $\mathcal{J}\uparrow$ & $\mathcal{F}\uparrow$\\
			\hline \hline
			Image-level Baseline & 40.9 & 40.5 & 41.3 \\
			\hline
			+Video-level Propagation $\mathcal{F}^{\text{Prop}}$ (Eq.~\ref{eq:1}) & 49.2 & 47.5 & 50.9 \\
			+Transformer-based Grounding $\mathcal{F}^{\text{Gnd}}$ (Eq.~\ref{eq:5}) & 56.4 & 54.8 & 58.1   \\
			+Sequence-NMS (Eq.~\ref{eq:2}) \& Model Ensemble & 61.4 & 60.0 & 62.7   \\
			\hline
		\end{tabular}
	}
	\captionsetup{font=small}
	\caption{\small\textbf{Ablation study} of essential components on \texttt{test-dev}.
	}
	\vspace{-14pt}
	\label{table:ablation}
\end{table}

\noindent\textbf{Training Detail.}
For $\mathcal{F}^{\text{Seg}}$, HTC is trained on COCO~\cite{lin2014microsoft} without finetuning. CondInst is pretrained on COCO and finetuned on the training split of RVOS-D with the standard training setting in~\cite{tian2020conditional} for about $15,\!000$ steps.
The propagation module $\mathcal{F}^{\text{Prop}}$, \ie, CFBI+, is pretrained on COCO and finetuned over training split of VOS~\cite{xu2018youtube} track as a standard training setting in semi-supervised VOS task (see~\cite{yang2020collaborative} for more details).
For tracklet-language grounding module (Eqs.~\ref{eq:3}-~\ref{eq:5}), we pretrain it using the data from RefCOCO~\cite{yu2016modeling}, RefCOCOg~\cite{yu2016modeling} and RefCOCO+~\cite{mao2016generation} for about 20 epochs.
We use Adam~\cite{kingma2014adam} as the optimizer with a learning rate of 4e-5, batch size of $48$ and weight decay of 1e-4.
The module is further finetuned on the training split of RVOS-D for five epochs with a learning rate of 1e-5.

\noindent\textbf{Model Ensemble.}
Model ensemble is also used in our final submission. We build five models with different implementations of the visual encoder $\mathcal{E}^{\text{Visual}}$, \ie, ResNet101~\cite{he2016deep}, HRNet~\cite{sun2019deep} and ResNeSt101~\cite{zhang2020resnest}, and linguistic encoder $\mathcal{E}^{\text{Linguistic}}$, \ie, $\text{Deberta}_{\text{BASE}}$~\cite{he2020deberta} and $\text{Bart}_{\text{BASE}}$~\cite{lewis2019bart}. We use different hyperparameter settings to further promote model performance and simply average grounding probabilities from different models for final prediction.

\noindent\textbf{Results on RVOS Challenge.}
Table~\ref{table:dev} and Table~\ref{table:challenge} show the ranking result of top teams in \texttt{test-dev} and \texttt{test-challenge} sets respectively.
Our approach achieves the best performance on both the two sets across all the metrics and outperforms $2^{nd}$ best team with a large margin, \ie, $\textbf{11.3\%}$ in terms of overall $\mathcal{J}$\&$\mathcal{F}$ on \texttt{test-challenge}.
Fig.~\ref{fig:quali} shows qualitative results of our proposed model on \texttt{test-challenge}.
With the effective top-down model design, our approach generates robust predictions even in challenging scenes, \eg, semantically similar instances, inconspicuous referent, complex linguistic description, \etc.

\noindent\textbf{Ablation Study.}
We start our ablation study with a simple image-level grounding pipeline ($1^{st}$ row in Table~\ref{table:ablation}), which only contains an image instance segmentation module $\mathcal{F}^{\text{Seg}}$ (Eq.~\ref{eq:0}) for image-level object candidates generation and implements the grounding module $\mathcal{F}^{\text{Gnd}}$ (Eq.~\ref{eq:5}) as a na\"{i}ve feature similarity operation without tracklet construction $\mathcal{F}^{\text{Prop}}$ (Eq.~\ref{eq:1}).
Then we progressively add essential modules ($2^{nd}$-$4^{th}$ rows in Table~\ref{table:ablation}).
With the fully exploration of intra- and inter-modal interactions, consistent performance improvements can be achieved.


{\small
\bibliographystyle{ieee_fullname}
\bibliography{egbib}
}

\end{document}